\documentclass{article}

\usepackage{arxiv}

\usepackage{lipsum}
\usepackage{graphicx}
\graphicspath{ {./images/} }

\usepackage[utf8]{inputenc}
\usepackage[T1]{fontenc}
\usepackage{hyperref}
\usepackage{url}
\usepackage{booktabs}
\usepackage{amsmath,amssymb,amsfonts,amsthm,mathtools}
\usepackage{bm}
\usepackage{nicefrac}
\usepackage{microtype}
\usepackage{xcolor}
\usepackage{graphicx}
\usepackage{algorithm}
\usepackage{algpseudocode}
\usepackage{multirow}
\usepackage{makecell}
\usepackage{bbm}
\newtheorem{theorem}{Theorem}[section]
\newtheorem{proposition}[theorem]{Proposition}

\usepackage{dsfont}
\newcommand{\ours}{CFHRL}
\newcommand{\D}{\mathcal{D}}
\newcommand{\G}{\mathcal{G}}
\newcommand{\Sspace}{\mathcal{S}}
\newcommand{\Aspace}{\mathcal{A}}
\newcommand{\E}{\mathbb{E}}

\usepackage{natbib}
\newcommand{\Dphi}{D_{\phi}}
\newcommand{\Vphi}{V_{\phi}}
\newcommand{\piplan}{\pi_{\mathrm{p}}}
\newcommand{\piz}{\pi_{\mathrm{z}}}
\newcommand{\pia}{\pi_{\mathrm{a}}}
\newcommand{\gbar}{\bar{g}}

\title{Adaptive Coarse-to-Fine Subgoal Refinement for Long-Horizon Offline Goal-Conditioned Reinforcement Learning}

\author{
\begin{tabular}{c}
Kaiqiang Ke$^{1,2}$ \quad
Shenghong He$^{1}$ \quad
Chengdong Xu$^{1,2}$ \quad
Yuheng Luo$^{1}$ \quad
Xiangyuan Lan$^{2}$ \quad
Chao Yu$^{1}$ \\
\\
$^{1}$Sun Yat-sen University \\
$^{2}$Pengcheng Laboratory \\
\\
\texttt{\{kekq,heshh23,xuchd6,luoyh227\}@mail2.sysu.edu.cn} \\
\texttt{lanxy@pcl.ac.cn} \quad
\texttt{yuchao3@mail.sysu.edu.cn}
\end{tabular}
}

\begin{document}

\maketitle

\begin{abstract}
Offline goal-conditioned reinforcement learning (GCRL) is challenging in long-horizon tasks, where distant state--goal pairs provide weak supervision and value estimates become vulnerable to accumulated bootstrapping errors.
Hierarchical methods mitigate this difficulty by introducing intermediate subgoals, but fixed temporal abstractions or fixed hierarchy depths can be mismatched to state--goal pairs with different reachability horizons.
We propose \textbf{Coarse-to-Fine Hierarchical Goal Reinforcement Learning} (\ours), a fully offline GCRL framework that adaptively refines distant goals before execution.
Starting from the final goal, \ours{} recursively proposes intermediate targets, trained from replay-supported candidates, and stops refinement once the current target is estimated to be locally executable by a learned reachability cost.
The key idea is that a subgoal need not be an exact midpoint or globally optimal waypoint; it only needs to provide reliable progress and reduce the remaining reaching difficulty, enabling subsequent refinement over shorter horizons.
A stylized analysis further supports the robustness of approximate recursive contraction.
Experiments on OGBench show substantial gains on several long-horizon tasks, with ablations validating the proposed refinement and stopping mechanisms.
\end{abstract}

\section{Introduction}

Offline goal-conditioned reinforcement learning (GCRL) aims to learn policies that reach diverse goals from a fixed dataset of previously collected trajectories~\citep{schaul2015universal,andrychowicz2017hindsight,eysenbach2022contrastive,park2024ogbench}.
This setting is appealing for robotics and embodied-control problems, where online exploration may be expensive, unsafe, or impractical~\citep{li2022hierarchical,giammarino2025physics}.
However, offline GCRL becomes particularly challenging when the desired goal is far from the current state.
This difficulty mainly arises from long-horizon value estimation: under sparse goal-reaching rewards, distant state--goal pairs provide little direct supervision, and the value function has to propagate goal information through many bootstrapped updates~\citep{kostrikov2022offline,park2025transitive,ni2025longhorizon}.
As the temporal distance increases, approximation errors can accumulate along the backup chain and lead to inaccurate evaluations of faraway goals~\citep{sutton1999between,nachum2018data,levy2019learning,park2023hiql,ahn2025option}.
Consequently, directly training a flat goal-conditioned policy from this value function can be unreliable, as the policy must make long-horizon decisions based on potentially inaccurate value estimates.

\begin{figure}[t]
    \centering
    \includegraphics[width=0.96\linewidth]{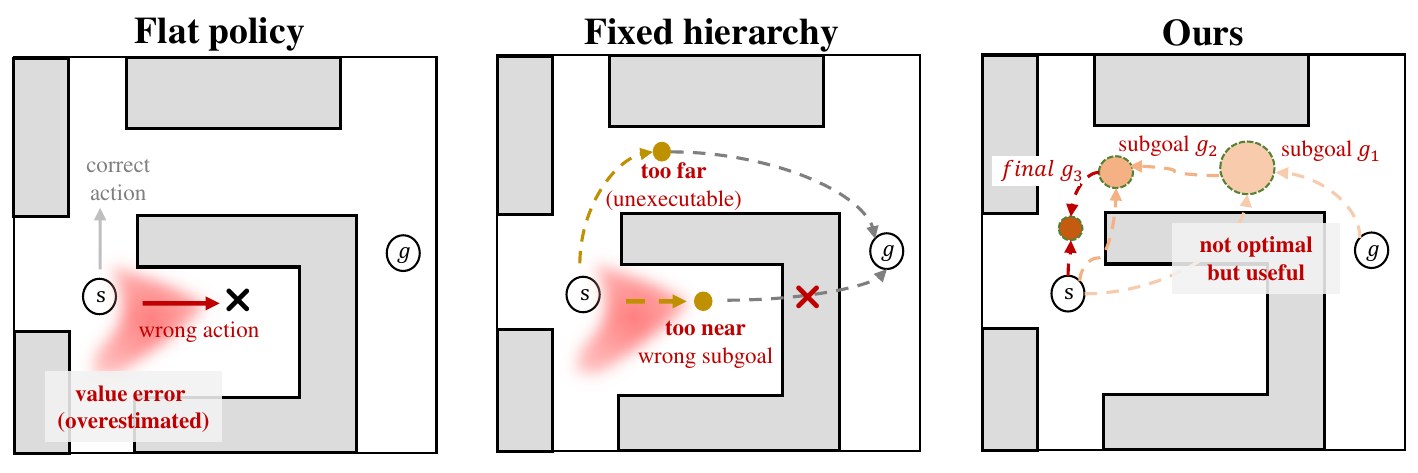}
    \caption{
    Motivation of adaptive stopping-time coarse-to-fine refinement.
    A flat goal-conditioned policy directly depends on noisy long-horizon value estimates.
    A fixed hierarchy introduces subgoals but uses a prescribed temporal scale or a fixed number of high-level decisions, which can be mismatched to state-goal pairs with different reachability horizons.
    \ours{} instead refines the current planning target only when it is estimated to be outside the executable region of the low-level policy, and stops once the target becomes locally reachable.
    }
    \label{fig:motivation}
    \vspace{-0.5em}
\end{figure}

Hierarchical decomposition provides a natural way to mitigate the long-horizon difficulty in offline GCRL.
Rather than requiring a flat policy to directly reach a distant goal, hierarchical methods introduce intermediate subgoals so that long-horizon goal reaching can be reduced to a sequence of shorter-range decisions~\citep{sutton1999between,nachum2018data,levy2019learning,park2023hiql,ahn2025option}.
Existing methods typically instantiate this idea through fixed temporal abstractions, trajectory-based subgoal extraction, or learned high-level subgoal policies~\citep{chanesane2021goal,park2023hiql,ahn2025option}.
Despite their effectiveness, these methods often rely on a prescribed decomposition scale, such as a fixed subgoal horizon or a fixed number of high-level decisions.
Such a fixed scale can be mismatched to state--goal pairs with different
reachability horizons: as shown in Figure~\ref{fig:motivation}, overly local
subgoals may make only marginal progress toward the final goal, so their small
value differences can be dominated by long-horizon estimation errors and lead to
misleading subgoal choices; whereas overly distant subgoals may fall outside the
reliable execution range of the low-level policy.

This raises a natural question: \emph{can an offline hierarchical agent adaptively identify a suitable intermediate target for each state--goal pair?}
Our key idea is to solve long-horizon goal reaching in a coarse-to-fine manner.
Although learned value functions can be noisy, coarse refinement does not require fine-grained discrimination among nearby states.
At the initial stage, the candidate targets differ at a larger scale, so the value function only needs to provide a directional signal toward a replay-supported region that reduces the remaining reachability horizon.
The selected subgoal therefore need not be optimal; it only needs to make meaningful progress and enable subsequent, finer refinement over shorter and more reliable horizons.
As illustrated in the right panel of Figure~\ref{fig:motivation}, this adaptive process converts a distant goal-reaching problem into a sequence of locally manageable subproblems.

Motivated by this insight, we propose \textbf{Coarse-to-Fine Hierarchical Goal Reinforcement Learning} (\ours), a fully offline GCRL framework that adaptively refines distant goals before execution.
Starting from the final goal, \ours{} recursively replaces the current planning target with intermediate targets produced by a planner trained on replay-supported candidates, until the target is estimated to be locally executable by a learned reachability cost.
The refinement planner is trained from replay-supported candidates, using a reachability-based objective that favors intermediate targets which reduce the bottleneck cost between the current state and the original goal.
After refinement, the selected target is executed through an abstraction policy and a low-level goal-conditioned policy, both trained by progress-weighted imitation over short-horizon offline segments.
Overall, \ours{} uses a shared reachability signal to coordinate target selection, adaptive stopping, and local execution, allowing the effective decomposition depth to be determined by the estimated reachability of each state--goal pair in a fully offline manner.
Our contributions are threefold:
(1) We formulate long-horizon offline GCRL as adaptive stopping-time subgoal refinement, where the effective decomposition depth is determined by learned executability rather than a fixed horizon.
(2) We introduce a replay-supported contraction objective that trains an amortized refinement planner without test-time graph search.
(3) We provide a stylized motivating analysis and empirical ablations illustrating when approximate contraction and adaptive stopping can improve long-horizon maze-style goal reaching.

\section{Preliminaries}
\label{sec:preliminaries}

\subsection{Offline Goal-Conditioned Reinforcement Learning}

We consider an offline goal-conditioned Markov decision process
$\mathcal{M}=(\Sspace,\Aspace,\G,P,r,\gamma,\rho_0)$, where $\Sspace$ and $\Aspace$ denote the state and action spaces, $\G$ denotes the goal space, $P$ is the transition dynamics, $\gamma\in[0,1)$ is the discount factor, and $\rho_0$ is the initial-state distribution.
In this work, goals share the same representation space as states; we use $s$ and $g$ only to distinguish the current state from the desired target.

The objective is to learn a goal-conditioned policy $\pi(a\mid s,g)$ from a fixed offline dataset $\D=\{\tau_i\}_{i=1}^N$, where $\tau_i=(s_0^i,a_0^i,s_1^i,\ldots,s_{T_i}^i)$, without further environment interaction.
Following the standard sparse goal-reaching formulation, the reward is defined as
\[
r(s_t,a_t,g)=
\begin{cases}
0, & \text{if } s_{t+1} \text{ reaches } g,\\
-1, & \text{otherwise}.
\end{cases}
\]
Thus, values are non-positive, and values closer to zero indicate shorter expected reaching time.
The main difficulty in long-horizon offline GCRL is that supervision for distant goals must be propagated through many transitions using only fixed offline data.

\subsection{Goal-Conditioned Implicit Value Learning}

CFHRL uses a unified goal-conditioned value function as the reachability estimator.
Following GCIVL~\citep{park2024ogbench}, we learn a $V$-only goal-conditioned value function from hindsight-relabeled offline transitions.
Let $(s,s')\sim\D$ denote a state transition from the offline dataset, and let $g\sim p_{\mathrm{mix}}^{\D}(\cdot\mid s)$ be a relabeled goal sampled from a mixture of current, future, and random goals.
The value function is trained by the expectile objective
\begin{equation}
\mathcal{L}_{\mathrm{GCIVL}}(\phi)
=
\E_{(s,s')\sim\D,\; g\sim p_{\mathrm{mix}}^{\D}(\cdot\mid s)}
\left[
\ell_2^\tau
\left(
r_g(s) + \gamma \bar V(s',g) - \Vphi(s,g)
\right)
\right],
\label{eq:gcivl}
\end{equation}
where $\bar V$ is the target value network, $r_g$ follows the sparse goal-reaching reward defined above, and
$\ell_2^\tau(u)=|\tau-\mathds{1}\{u<0\}|u^2$ is the expectile regression loss.

We convert the learned value into a non-negative reachability cost by
\begin{equation}
\Dphi(s,g) = \left[-\widehat V_\phi(s,g)\right]_+,
\label{eq:reachability-cost}
\end{equation}
where $[x]_+=\max(x,0)$. For a single value network, $\widehat V_\phi=V_\phi$; when an ensemble is used, we take the conservative estimate $\widehat V_\phi(s,g)=\min_m V_{\phi_m}(s,g)$.
Since sparse goal-reaching values are non-positive in the ideal case, this reduces to $\Dphi(s,g)=-\Vphi(s,g)$ up to numerical clipping.
Thus, smaller $\Dphi(s,g)$ indicates that $g$ is estimated to be more reachable from $s$.

\subsection{Midpoint-Based Subgoal Learning}

A common strategy for long-horizon goal reaching is to introduce an intermediate subgoal $g'$ between the current state $s$ and final goal $g$~\citep{jurgenson2020sub,Gong2024automatic,chanesane2021goal,subgoal2024diffuser}.
Given a reachability distance $d$, a midpoint-style objective evaluates a candidate subgoal by
\begin{equation}
C(g'\mid s,g)=\max\{d(s,g'),d(g',g)\}.
\label{eq:midpoint_cost}
\end{equation}
Minimizing this cost encourages $g'$ to balance the two subproblems $s\rightarrow g'$ and $g'\rightarrow g$.
This perspective is important but incomplete for heterogeneous horizons.
A one-shot subgoal can still be too distant for local execution, and a fixed hierarchy depth may be insufficient as the state-goal distance grows.
\ours{} extends the midpoint principle into a recursive procedure: it repeatedly selects value-guided contracting subgoals until the remaining target is estimated to be executable by the local controller.

\section{Method}

\subsection{Overview}

CFHRL is an adaptive hierarchical framework for long-horizon offline goal-conditioned reinforcement learning. Given a current state $s$ and a final goal $g$, CFHRL first refines the distant goal into a locally executable target, then converts this target into an execution-oriented command, and finally uses a low-level policy to produce primitive actions.

The framework consists of three learned policies. The refinement planner $\piplan(u\mid s,g)$ proposes an intermediate target $u$ in the same state--goal representation used by the reachability model, conditioned on the current state and the current planning target. The abstraction policy $\piz(z\mid s,\gbar)$ maps the refined target $\gbar$ to a local command $z$. The low-level policy $\pia(a\mid s,z)$ maps the command to a primitive action. These policies are coordinated by the shared reachability cost $\Dphi(s,g)$ in Eq.~\eqref{eq:reachability-cost}. The same cost is used for value-guided planner training, executability checking, and progress-weighted local imitation.

\subsection{Adaptive Coarse-to-Fine Refinement}

CFHRL replaces a fixed subgoal scale with an adaptive refinement process.
Given the current state $s$ and the final goal $g$, the planner initializes the planning target as
$g^{(0)}=g$ and progressively refines it into a more locally reachable target.
At refinement level $\ell$, the current target $g^{(\ell)}$ is regarded as locally executable from $s$ if
\begin{equation}
D_\phi(s,g^{(\ell)}) \leq \epsilon_{\rm exec},
\label{eq:stopping_condition}
\end{equation}
where $\epsilon_{\rm exec}$ is the executability threshold. It specifies the maximum estimated
reachability cost under which the target is directly passed to the execution policies.
If Eq.~\eqref{eq:stopping_condition} is satisfied, refinement stops and $g^{(\ell)}$ is passed to the execution module.
Otherwise, the planner proposes an intermediate target
$u^{(\ell)}\sim\piplan(\cdot\mid s,g^{(\ell)})$, and the planning target is updated as
\[
g^{(\ell+1)} = u^{(\ell)} .
\]

This procedure is repeated until the target becomes locally executable or the maximum refinement depth $K_{\max}$ is reached.
The resulting stopping level is
\begin{equation}
\ell^\star(s,g)
=
\min\left\{
\ell\in\{0,\ldots,K_{\max}\}:
D_\phi(s,g^{(\ell)})\leq \epsilon_{\rm exec}
\right\},
\label{eq:stopping_level}
\end{equation}
with $\ell^\star(s,g)=K_{\max}$ if no target satisfies the stopping condition.
The final refined target is then
\begin{equation}
\gbar = g^{(\ell^\star(s,g))}.
\label{eq:refined-target}
\end{equation}

Thus, the refinement depth is determined by estimated reachability rather than a fixed temporal scale~\citep{ahn2025option,hwang2024strict,temporal2025subgoal}.
Nearby goals can be executed directly, while distant goals are recursively replaced by intermediate targets.
The planner need not predict an exact midpoint; it only needs to propose targets that reduce the remaining reaching difficulty.

\subsection{Offline Learning of the Refinement Planner}

The refinement planner is trained entirely from offline data. Given a current observation $s_t$
and a high-level target $g$, we construct a replay-supported candidate set
\begin{equation}
\mathcal C_D(s_t,g)=\{\tilde s_i\}_{i=1}^{N_c},
\label{eq:candidate_set}
\end{equation}
where each $\tilde s_i$ is a full observation sampled from the offline dataset and used as a candidate target for planner supervision~\citep{eysenbach2019search,graph2025stitching}.
When $g$ comes from the same trajectory as $s_t$, candidates are sampled from intermediate future states between $s_t$ and $g$; otherwise, they are sampled from the replay state distribution.
This construction keeps subgoal supervision close to the data support.

Each candidate is scored by the bottleneck reachability cost of the two induced subproblems:
\begin{equation}
C_\phi(\tilde s_i\mid s_t,g)
=
\max\{D_\phi(s_t,\tilde s_i), D_\phi(\tilde s_i,g)\}.
\label{eq:decomposition_cost}
\end{equation}
Both terms query the same reachability model with full-observation state--target pairs.
The first term estimates the cost of reaching the candidate observation $\tilde s_i$ from $s_t$,
while the second estimates the remaining cost of reaching $g$ from $\tilde s_i$.
A lower score indicates that the candidate better contracts the original reaching problem.

We train the planner by value-guided weighted imitation.
The candidate weights are defined as
\begin{equation}
w_i
=
\frac{
\exp\left(-C_{\phi}(\tilde{s}_i\mid s_t,g)/\eta_{\mathrm{p}}\right)
}{
\sum_{j=1}^{N_c}
\exp\left(-C_{\phi}(\tilde{s}_j\mid s_t,g)/\eta_{\mathrm{p}}\right)
},
\label{eq:planner-weight}
\end{equation}
where $\eta_{\mathrm{p}}$ is the planner temperature.
The planner is optimized with
\begin{equation}
\mathcal{L}_p
=
-
\mathbb{E}_{(s_t,g,\mathcal{C}_{\mathcal{D}})}
\left[
\sum_{i=1}^{N_c}
\mathrm{sg}(w_i)
\log \pi_p(\tilde s_i\mid s_t,g)
\right].
\label{eq:planner-loss}
\end{equation}
Thus, the planner learns to imitate replay-supported subgoals that decompose a distant goal into easier residual reaching problems, without requiring online interaction or manually labeled subgoals. At inference time, the planner samples directly from this learned policy; replay support is therefore imposed through offline supervision rather than by explicit nearest-neighbor projection or test-time replay-buffer search.

\subsection{Learning the Execution Policies}

After refinement, CFHRL executes the selected target $\gbar$ through the same abstraction interface as HIQL~\citep{park2023hiql}.
The abstraction policy $\piz(z\mid s,\gbar)$ converts the refined target into a local command $z$, and the low-level policy $\pia(a\mid s,z)$ maps this command to a primitive action.

To train these policies, we sample short-horizon segments $(s_t,a_t,s_{t+h})$ from the offline dataset, where $h$ lies within a local execution window.
The future state $s_{t+h}$ serves as the local target, and the corresponding command is constructed as $z_t=\psi(s_t,s_{t+h})$.
Here $\psi$ denotes the command-construction map used by the low-level actor. In the default setting, learned policy-side goal representations are disabled and $z_t$ is the raw local goal representation. When learned representations are enabled, $\psi$ is the goal encoder trained with the value function and detached when used as an actor target.
The abstraction policy learns to predict $z_t$ from $(s_t,s_{t+h})$, while the low-level policy imitates the dataset action $a_t$ conditioned on $(s_t,z_t)$.

Both policies are trained by value-improvement-weighted behavior cloning~\citep{ghosh2021learning,ma2022gofar,zeng2025sihd}.
For the local target $s_{t+h}$, we define
\begin{equation}
\omega_t
=
\min\left\{
\exp\left(
\alpha\left[
\Vphi(s_{t+1},s_{t+h})-\Vphi(s_t,s_{t+h})
\right]
\right),
\omega_{\max}
\right\}.
\label{eq:progress-weight}
\end{equation}
The execution losses are
\begin{align}
\mathcal{L}_{\mathrm{z}}
&=
-
\E_{\D}
\left[
\omega_t
\log \piz(z_t\mid s_t,s_{t+h})
\right],
\\
\mathcal{L}_{\mathrm{a}}
&=
-
\E_{\D}
\left[
\omega_t
\log \pia(a_t\mid s_t,z_t)
\right].
\end{align}
The weight $\omega_t$ assigns larger imitation weight to transitions that increase the learned value toward the local target, thereby biasing both execution policies toward locally progressive behavior.

At test time, CFHRL refines the final goal into $\gbar$, samples $z\sim\piz(\cdot\mid s,\gbar)$, and executes $\pia(\cdot\mid s,z)$ for a short interval before replanning.
This enables adaptive execution without global replay-buffer search at inference time.
The complete training and inference procedures are provided in Algorithms~\ref{alg:cfhrl-training}
and~\ref{alg:cfhrl-inference} in Appendix.

\section{A Motivating Analysis}
\label{sec:analysis}

This section provides an illustrative analysis of why adaptive coarse-to-fine refinement can be more robust than fixed local-scale subgoal selection under noisy long-horizon distance estimates.
The analysis is not intended as a general guarantee for arbitrary offline GCRL problems.
Instead, it highlights a key intuition: fixed local-scale decisions can still be
scored by noisy long-horizon residuals, whereas recursive refinement only requires
approximate contraction of the remaining reaching problem.
In CFHRL, the learned distance estimate $\widehat D$ corresponds to the reachability cost $\Dphi=-\Vphi$.

Let $D^\star(s,g)$ denote the true reachability distance and $\widehat D(s,g)$ its learned estimate.
Consider a one-dimensional problem with current state $s=0$ and a final goal $g^\star$ located at distance $T>0$ from the current state, so that $D^\star(x,g^\star)=|T-x|$.
Assume the learned estimate satisfies
\begin{equation}
\widehat D(x,g^\star)
=
D^\star(x,g^\star)+\xi(x,g^\star),
\qquad
\operatorname{Var}[\xi(x,g^\star)]
=
\sigma^2\big(D^\star(x,g^\star)\big)^{2\alpha},
\label{eq:noisy_distance_model}
\end{equation}
for $\alpha\in(0,1]$.
Here, $\sigma$ controls the overall magnitude of estimation noise, while
$\alpha$ captures that distance estimates become noisier as the residual horizon increases.

\begin{proposition}[Fixed local-scale decisions can have vanishing signal-to-noise]
\label{prop:fixed_snr}
Suppose a fixed local-scale hierarchy compares $x_+=k$ and $x_-=-k$, where $0<k<T$, $x_+$ moves toward the goal, and $x_-$ moves away from it.
Assume that the noise terms $\xi(x_+,g^\star)$ and $\xi(x_-,g^\star)$ are independent zero-mean Gaussian variables with variances specified by Eq.~\eqref{eq:noisy_distance_model}.
Then the probability of selecting the wrong candidate is
\begin{equation}
P_{\mathrm{fix}}(T)
=
\Phi\left(
-
\frac{2k}{\sigma\sqrt{(T-k)^{2\alpha}+(T+k)^{2\alpha}}}
\right),
\qquad
\lim_{T\rightarrow\infty}P_{\mathrm{fix}}(T)=\frac{1}{2},
\end{equation}
where $\Phi$ is the standard normal cumulative distribution function.
\end{proposition}

Proposition~\ref{prop:fixed_snr} shows that choosing a fixed local subgoal scale does not necessarily remove the long-horizon estimation burden.
Although the candidate step size is fixed, the compared residual distances remain of order $T$, while the useful decision margin is only $O(k)$.

We next analyze value-guided contraction.
For a candidate subgoal $x$, define the true decomposition cost as
\begin{equation}
C^\star(x\mid s,g)
=
\max\{D^\star(s,x),D^\star(x,g)\}.
\end{equation}
For $\rho\in(1/2,1)$, define the $\rho$-contraction set
\begin{equation}
\G_\rho(s,g)
=
\{x:C^\star(x\mid s,g)\le \rho D^\star(s,g)\}.
\end{equation}
Any candidate in this set reduces the maximum residual horizon by a constant factor.

\begin{proposition}[Approximate contraction is sufficient]
\label{prop:contraction}
Let $T=D^\star(s,g^\star)$ and suppose the candidate set $\mathcal{X}$ contains a good candidate $x_g$ satisfying
$C^\star(x_g\mid s,g^\star)\le\rho_0T$, where $1/2\le\rho_0<\rho<1$.
Let
\begin{equation}
\widehat C(x\mid s,g^\star)
=
\max\{\widehat D(s,x),\widehat D(x,g^\star)\}.
\end{equation}
On the event
\begin{equation}
\mathcal{E}_\varepsilon
=
\left\{
\sup_{x\in\mathcal{X}}
\left|
\widehat C(x\mid s,g^\star)-C^\star(x\mid s,g^\star)
\right|
\le \varepsilon T
\right\},
\end{equation}
if $\varepsilon<(\rho-\rho_0)/2$, then any minimizer
$\widehat x=\arg\min_{x\in\mathcal{X}}\widehat C(x\mid s,g^\star)$
belongs to $\G_\rho(s,g^\star)$.
\end{proposition}

Proposition~\ref{prop:contraction} formalizes the weak requirement used by \ours{}: exact midpoint prediction is unnecessary.
It is enough for the learned cost to preserve a coarse margin between candidates that contract the horizon and those that do not.
The same argument also tolerates an approximate selector.
If a learned planner selects $\tilde x$ such that
$\widehat C(\tilde x\mid s,g^\star)\le \min_{x\in\mathcal{X}}\widehat C(x\mid s,g^\star)+\beta T$,
then $\tilde x\in\G_\rho(s,g^\star)$ whenever
$\varepsilon<(\rho-\rho_0-\beta)/2$.
Thus, the analysis characterizes the target behavior induced by value-guided planner training, rather than providing a finite-sample guarantee for the learned actor.

If each refinement step selects a $\rho$-contraction, the residual horizon after $L$ refinements satisfies $T_L\le\rho^LT_0$, so reaching a local executable radius $r$ requires only $O(\log(T_0/r)/\log(1/\rho))$ refinements.
In practice, refinement can fail and local execution is imperfect.
Let $\delta_\rho(T_\ell)$ be the probability of failing to select a $\rho$-contraction at residual horizon $T_\ell$, and let $\psi(T)$ upper-bound local execution error for a target with residual horizon $T$.
Conditioned on all $L$ refinement steps being successful, the remaining horizon is at most $\rho^LT_0$.
A union bound over refinement failures and final local execution gives
\begin{equation}
P_{\mathrm{err}}(L)
\le
\sum_{\ell=0}^{L-1}\delta_\rho(T_\ell)+\psi(\rho^L T_0).
\end{equation}
The term $\delta_\rho(T_\ell)$ summarizes all failures of the learned refinement step.
In Appendix~\ref{app:coverage-planner}, we further decompose this term into three concrete sources: insufficient replay coverage of contracting candidates, error in the learned reachability cost, and approximation error of the amortized planner.
This extended analysis connects the contraction view to the offline candidate budget used in CFHRL, while keeping the main text focused on the motivating mechanism.

\section{Experiments}
\label{sec:experiments}

We evaluate \ours{} on OGBench offline goal-conditioned control tasks~\citep{park2024ogbench}.
Our experiments assess whether adaptive coarse-to-fine refinement improves long-horizon goal reaching, verify the contributions of adaptive stopping and reachability-guided subgoal selection, and study the effect of the offline candidate budget used for planner supervision.

\subsection{Experimental Setup}

\paragraph{Environments.}
We evaluate on OGBench locomotion and manipulation tasks.
The locomotion suite includes PointMaze, AntMaze, HumanoidMaze, and AntSoccer, where agents must reach target goals through continuous-control navigation.
The manipulation suite includes Cube and Scene, where agents interact with movable objects to achieve specified goal configurations.
Together, these tasks cover diverse offline goal-conditioned control problems across navigation and manipulation domains.

\paragraph{Baselines.}
We compare with representative offline GCRL methods, including goal-conditioned behavior cloning (GCBC)~\citep{ding2019goal}, goal-conditioned implicit value learning (GCIVL)~\citep{park2024ogbench}, goal-conditioned implicit Q-learning (GCIQL)~\citep{park2024ogbench}, quasimetric reinforcement learning (QRL)~\citep{wang2023optimal}, contrastive reinforcement learning (CRL)~\citep{eysenbach2022contrastive}, and hierarchical implicit Q-learning (HIQL)~\citep{ghosh2021learning, eysenbach2022contrastive, park2023hiql, myers2025offline}.
HIQL is the most direct hierarchical baseline, as it also derives a hierarchical policy from a learned goal-conditioned value function.
We further evaluate controlled variants of \ours{} in Section~\ref{sec:ablations}. Unless otherwise stated, all baseline results use the official OGBench hyperparameters and tuning protocol.

\paragraph{Evaluation protocol.}
Each method is trained with $8$ random seeds and evaluated on $50$ episodes per task and seed.
We report mean success rates and standard deviations across seeds.
All methods use the same offline datasets and standard OGBench goal configurations.
The aggregate score is the unweighted average over the tasks reported in Table~\ref{tab:main_results}.


\subsection{Main Results}
\label{sec:main_results}

Table~\ref{tab:main_results} reports the main results on OGBench locomotion and manipulation tasks.
Across the reported tasks, \ours{} achieves the highest aggregate score, improving over HIQL from $23.2$ to $37.5$.
The improvement is mainly concentrated in long-horizon maze-style tasks, where the agent must compose multiple locally feasible transitions to reach a distant goal.
A representative example is PointMaze-Giant-Stitch, where \ours{} improves the success rate from $0\pm0$ to $57\pm8$.
This task is particularly challenging for flat or fixed-hierarchy methods because the target often cannot be reached through a single reliable value-guided decision.
By recursively refining the goal and stopping once the target becomes locally executable, \ours{} reduces the burden on both long-horizon value estimation and low-level control.

\begin{table*}[t]
\centering
\caption{
Main results on OGBench locomotion and manipulation tasks.
We report mean success rate and standard deviation over $8$ random seeds.
Best results in each row are shown in bold. The aggregate score is computed over the tasks shown in the table.
}
\label{tab:main_results}
\resizebox{\textwidth}{!}{
\begin{tabular}{llccccccc}
\toprule
\textbf{Environment} & \textbf{Dataset}
& \textbf{GCBC}
& \textbf{GCIVL}
& \textbf{GCIQL}
& \textbf{QRL}
& \textbf{CRL}
& \textbf{HIQL}
& \textbf{\ours} \\
\midrule
\multirow{2}{*}{PointMaze}
& pointmaze-giant-navigate-v0
& $1 \pm 2$ & $0 \pm 0$ & $0 \pm 0$ & $68 \pm 7$ & $27 \pm 10$ & $46 \pm 9$ & $\mathbf{82 \pm 11}$ \\
& pointmaze-giant-stitch-v0
& $0 \pm 0$ & $0 \pm 0$ & $0 \pm 0$ & $50 \pm 8$ & $0 \pm 0$ & $0 \pm 0$ & $\mathbf{57 \pm 8}$ \\
\midrule
\multirow{2}{*}{AntMaze}
& antmaze-giant-navigate-v0
& $0 \pm 0$ & $0 \pm 0$ & $0 \pm 0$ & $14 \pm 3$ & $16 \pm 3$ & $65 \pm 5$ & $\mathbf{68 \pm 4}$ \\
& antmaze-giant-stitch-v0
& $0 \pm 0$ & $0 \pm 0$ & $0 \pm 0$ & $0 \pm 0$ & $0 \pm 0$ & $2 \pm 2$ & $0 \pm 0$ \\
\midrule
\multirow{2}{*}{HumanoidMaze}
& humanoidmaze-giant-navigate-v0
& $0 \pm 0$ & $0 \pm 0$ & $0 \pm 0$ & $1 \pm 0$ & $3 \pm 2$ & $12 \pm 4$ & $\mathbf{42 \pm 6}$ \\
& humanoidmaze-giant-stitch-v0
& $0 \pm 0$ & $0 \pm 0$ & $0 \pm 0$ & $0 \pm 0$ & $0 \pm 0$ & $0 \pm 0$ & $0 \pm 0$ \\
\midrule
\multirow{2}{*}{AntSoccer}
& antsoccer-arena-navigate-v0
& $5 \pm 1$ & $47 \pm 3$ & $50 \pm 2$ & $8 \pm 2$ & $23 \pm 2$ & $\mathbf{58 \pm 2}$ & $\mathbf{58 \pm 3}$ \\
& antsoccer-medium-navigate-v0
& $2 \pm 0$ & $4 \pm 1$ & $7 \pm 1$ & $2 \pm 2$ & $3 \pm 1$ & $13 \pm 2$ & $\mathbf{16 \pm 2}$ \\
\midrule
\multirow{2}{*}{Cube}
& cube-single-play-v0
& $6 \pm 2$ & $53 \pm 4$ & $\mathbf{68 \pm 6}$ & $5 \pm 1$ & $19 \pm 2$ & $15 \pm 3$ & $23 \pm 2$ \\
& cube-double-play-v0
& $1 \pm 1$ & $36 \pm 3$ & $\mathbf{40 \pm 5}$ & $1 \pm 0$ & $10 \pm 2$ & $6 \pm 2$ & $4 \pm 2$ \\
\midrule
\multirow{1}{*}{Scene}
& scene-play-v0
& $5 \pm 1$ & $42 \pm 4$ & $51 \pm 4$ & $5 \pm 1$ & $19 \pm 2$ & $38 \pm 3$ & $\mathbf{62 \pm 5}$ \\
\midrule
\multicolumn{2}{l}{\textbf{Aggregate score}}
& $1.8$ & $16.5$ & $19.6$ & $14.0$ & $10.9$ & $23.2$ & $\mathbf{37.5}$ \\
\bottomrule
\end{tabular}
}
\end{table*}

The comparison with HIQL further suggests that the key advantage is not merely the use of hierarchy, but the ability to adapt the effective refinement depth to each state-goal pair.
Fixed two-level abstraction can be mismatched when the selected subgoal is either too distant for the low-level policy or too conservative to make meaningful progress.
In contrast, \ours{} uses the learned reachability cost to decide whether additional refinement is needed.
This mechanism is especially useful in PointMaze and HumanoidMaze, where different start-goal pairs can have substantially different effective horizons even within the same environment.
Qualitative planning visualizations in Appendix~\ref{app:planning_visualization} show that the learned planner first selects coarse intermediate targets and then progressively refines them into more local execution targets as the agent replans.

The results also reveal clear limitations of the proposed decomposition assumption.
On Cube tasks, flat value-based methods such as GCIQL outperform CFHRL, suggesting that these tasks are less naturally captured by replay-supported waypoint-like decomposition and are more sensitive to precise contact dynamics and object-state changes.
The strong performance on Scene-Play, however, indicates that adaptive refinement can still help manipulation-like domains when the task contains substantial repositioning or long-horizon goal decomposition.

\subsection{Controlled Ablations}
\label{sec:ablations}

Table~\ref{tab:ablations} evaluates the main design choices of \ours{} on representative long-horizon locomotion tasks.
All variants use the same value-learning backend, so the comparison isolates the effect of refinement and execution design.

First, the fixed-depth variants show that simply adding more forced refinement steps is not sufficient. In these variants, the planner performs exactly 
$K$ refinement steps without the adaptive stopping rule. The one-step variant improves over HIQL on PointMaze-Giant, but forcing the planner to refine for 
$K=2$ or $K=3$ substantially degrades performance. This indicates that excessive refinement can accumulate planner errors or produce targets that are no longer suitable for local execution. These results suggest that the benefit of CFHRL does not come from simply increasing the number of high-level decisions. In contrast, full CFHRL uses a maximum refinement depth together with value-thresholded stopping, so the actual number of refinement steps can be smaller than the maximum depth when the current target is already estimated to be locally executable.

\begin{table}[t]
\centering
\caption{Controlled ablations isolating the main design choices of CFHRL. We report success rates on representative long-horizon locomotion tasks.}
\label{tab:ablations}
\begin{tabular}{lccc}
\toprule
Variant & PointMaze-Giant & HumanoidMaze-Giant & AntSoccer-Medium \\
\midrule
Flat GCIVL & $0\pm0$ & $0\pm0$ & $4\pm1$ \\
CFHRL w/o planner & $46\pm9$ & $12\pm4$ & $13\pm2$ \\
Fixed depth $K=1$ & $69\pm8$ & $9\pm4$ & $16\pm2$ \\
Fixed depth $K=2$ & $37\pm 1$ & $2\pm1$ & $0\pm0$ \\
Fixed depth $K=3$ & $17\pm 4$ & $0\pm 0$ & $0\pm0$ \\
CFHRL w/o abstraction policy & $35\pm7$ & $16\pm5$ & $8\pm2$ \\
CFHRL w/ random dataset subgoals & $0\pm0$ & $0\pm0$ & $0\pm0$ \\
CFHRL full & $\mathbf{82\pm11}$ & $\mathbf{42\pm6}$ & $\mathbf{16\pm2}$ \\
\bottomrule
\end{tabular}
\end{table}

Second, the random-subgoal variant fails on all tasks, confirming that the improvement does not come from conditioning the policy on arbitrary intermediate states.
The subgoals must be selected according to the learned reachability cost so that they contract the remaining goal-reaching problem.
Removing the abstraction policy also reduces performance, showing that the refined target should be converted into an execution-oriented local command before being passed to the low-level controller.

Additional sensitivity analyses on the maximum planning steps and the stopping threshold $\epsilon$ are provided in Appendix~\ref{app:additional_ablations}.
\subsection{Training Candidate Count}
\label{sec:candidate_count}

Table~\ref{tab:candidate_budget} studies the effect of the training candidate budget $N_c$ used for value-guided planner supervision.
Increasing $N_c$ gives the planner access to a larger replay-supported candidate set, but also reduces training throughput because more candidate subgoals must be evaluated by the reachability model.
Here, mean refinement calls denotes the average number of planner refinement calls during evaluation.

The results show a clear trade-off.
Small candidate sets provide weak supervision and lead to lower performance, especially on HumanoidMaze-Giant.
Increasing $N_c$ from $1$ to $16$ improves success rates and reduces the number of refinement calls, indicating that the planner learns to propose more useful intermediate targets.
Further increasing $N_c$ to $32$ does not improve performance and substantially lowers the number of gradient updates per second.
We therefore use $N_c=16$ as the default training candidate budget in the main experiments.

\begin{table}[t]
\centering
\caption{
Effect of the training candidate budget $N_c$ on performance and training throughput.
The throughput is measured as the number of gradient updates per second on the same hardware.
Larger $N_c$ provides stronger value-guided supervision for the planner but increases training cost.
}
\label{tab:candidate_budget}
\begin{tabular}{lcccc}
\toprule
\textbf{$N_c$} 
& \textbf{PointMaze-Giant} 
& \textbf{HumanoidMaze-Giant} 
& \textbf{Updates / sec. $\uparrow$} 
& \textbf{Mean refinement calls} \\
\midrule
$1$  & $58$ & $12$ & $225$ & $3.8$ \\
$4$  & $74$ & $35$ & $201$ & $2.1$ \\
$8$  & $79$ & $39$ & $172$ & $1.8$ \\
$16$ & $82$ & $42$ & $119$ & $1.6$ \\
$32$ & $79$ & $39$ & $70$  & $1.7$ \\
\bottomrule
\end{tabular}
\end{table}

\section{Related Work}
\label{sec:related_work}

\paragraph{Offline goal-conditioned reinforcement learning.}
Goal-conditioned reinforcement learning learns policies that can reach goals specified at test time~\citep{schaul2015universal,andrychowicz2017hindsight}.
In offline GCRL, the agent learns from a fixed dataset and must generalize to new state-goal pairs without additional interaction~\citep{park2024ogbench}.
Goal-conditioned behavior cloning provides a simple baseline~\citep{lynch2019learning,ghosh2021learning}, while value-based methods such as GCIVL and GCIQL adapt implicit value learning to goal-reaching tasks~\citep{kostrikov2022offline,park2023hiql,park2024ogbench}.
Representation-based approaches such as contrastive RL and quasimetric RL learn structured goal-reaching similarities or distances~\citep{eysenbach2022contrastive,myers2025offline}.
These methods are strong baselines, but distant goals remain challenging because flat value propagation must bridge long temporal gaps directly.

\paragraph{Hierarchical and subgoal-based reinforcement learning.}
Hierarchical reinforcement learning uses temporal abstraction to solve long-horizon tasks through options, high-level controllers, or learned subgoals~\citep{sutton1999between,nachum2018data,levy2019learning}.
Subgoal-based GCRL learns intermediate targets that decompose difficult state-goal problems into shorter sub-problems~\citep{jurgenson2020sub,chanesane2021goal,Gong2024automatic}.
HIQL extracts a two-level hierarchical policy from a single goal-conditioned value function~\citep{park2023hiql}, while recent methods study option-aware temporally abstracted values and chain-of-goals policies~\citep{ahn2025option,choi2025chain}.
These approaches demonstrate the value of hierarchy, but they often rely on a fixed temporal abstraction or one-shot subgoal prediction.
CFHRL differs by recursively refining goals and using a learned executability threshold to adapt the effective planning depth to each state-goal pair.


\section{Conclusion}
\label{sec:conclusion}

We proposed \ours{}, an adaptive hierarchical offline GCRL framework for long-horizon goal reaching.
\ours{} recursively refines distant goals into more local targets and stops once the target is estimated to be executable.
Experiments on OGBench show strong gains on maze-style long-horizon tasks, and ablations validate the roles of recursive refinement, adaptive stopping, and reachability-guided candidate selection.
The weaker results on contact-rich manipulation suggest that future work should combine adaptive planning with richer object-centric abstractions.

\bibliographystyle{plainnat}
\bibliography{references}
\newpage
\appendix
\section{Proofs}
\label{app:proofs}

This appendix provides the proofs for the stylized analysis in Section~\ref{sec:analysis}.
The analysis is intended to isolate the mechanism behind adaptive coarse-to-fine refinement.
It studies an idealized value-guided selector that chooses candidates according to an estimated decomposition cost.
In the implemented algorithm, this selector is approximated by a learned high-level actor trained from offline candidate subgoals.
Therefore, the results should be interpreted as mechanism-level justification rather than a finite-sample guarantee for the learned policy.

\subsection{Proof of Proposition~\ref{prop:fixed_snr}}
\label{app:proof_fixed}

We compare two fixed local-scale candidates in the one-dimensional problem.
The current state is $s=0$, and the final goal $g^\star$ is located at coordinate $T>0$.
Let $x_+=k$ denote the candidate that moves toward the goal and $x_-=-k$ denote the candidate that moves away from the goal, where $0<k<T$.
The true residual distances to the final goal are
\[
    D^\star(x_+,g^\star)=T-k,
    \qquad
    D^\star(x_-,g^\star)=T+k .
\]
Thus, the true margin between the incorrect and correct candidates is $2k$.

The learned distance estimate is
\[
    \widehat D(x,g^\star)
    =
    D^\star(x,g^\star)+\xi(x,g^\star),
\]
where the noise variance scales with the residual horizon as
\[
    \mathrm{Var}[\xi(x,g^\star)]
    =
    \sigma^2\big(D^\star(x,g^\star)\big)^{2\alpha}.
\]
The fixed local-scale selector makes an error when the estimated distance of the correct candidate is larger than that of the incorrect one:
\[
    \widehat D(x_+,g^\star)>\widehat D(x_-,g^\star).
\]
Substituting the noisy estimates gives
\[
    (T-k)+\xi(x_+,g^\star)
    >
    (T+k)+\xi(x_-,g^\star),
\]
or equivalently,
\[
    \xi(x_+,g^\star)-\xi(x_-,g^\star)>2k .
\]

By assumption, $\xi(x_+,g^\star)$ and $\xi(x_-,g^\star)$ are independent zero-mean Gaussian variables.
Therefore, the difference
$\xi(x_+,g^\star)-\xi(x_-,g^\star)$ is Gaussian with mean zero and variance
\[
    \sigma^2\left[(T-k)^{2\alpha}+(T+k)^{2\alpha}\right].
\]
It follows that the error probability is
\[
    P_{\mathrm{fix}}(T)
    =
    \Phi\left(
    -
    \frac{2k}{
    \sigma\sqrt{(T-k)^{2\alpha}+(T+k)^{2\alpha}}}
    \right),
\]
where $\Phi$ is the standard normal cumulative distribution function.
For any $\alpha>0$, the denominator diverges as $T\rightarrow\infty$, and hence the argument of $\Phi$ converges to zero.
Consequently,
\[
    \lim_{T\rightarrow\infty}P_{\mathrm{fix}}(T)=\frac{1}{2}.
\]

\subsection{Proof of Proposition~\ref{prop:contraction}}
\label{app:proof_contraction}

Let $T=D^\star(s,g^\star)$ be the current residual horizon.
Suppose the candidate set $\mathcal{X}$ contains a good candidate $x_g$ satisfying
\[
    C^\star(x_g\mid s,g^\star)\leq \rho_0 T,
\]
where $1/2\leq \rho_0<\rho<1$.
On the event $\mathcal{E}_\varepsilon$, the estimated decomposition cost satisfies
\[
    \left|
    \widehat C(x\mid s,g^\star)-C^\star(x\mid s,g^\star)
    \right|
    \leq \varepsilon T
\]
for all $x\in\mathcal{X}$.

By the error bound, the estimated cost of $x_g$ is bounded by
\[
    \widehat C(x_g\mid s,g^\star)
    \leq
    C^\star(x_g\mid s,g^\star)+\varepsilon T
    \leq
    (\rho_0+\varepsilon)T .
\]
Let $\widehat x$ be a minimizer of the estimated decomposition cost over $\mathcal{X}$.
Then
\[
    \widehat C(\widehat x\mid s,g^\star)
    \leq
    \widehat C(x_g\mid s,g^\star)
    \leq
    (\rho_0+\varepsilon)T .
\]
Applying the error bound again yields
\[
    C^\star(\widehat x\mid s,g^\star)
    \leq
    \widehat C(\widehat x\mid s,g^\star)+\varepsilon T
    \leq
    (\rho_0+2\varepsilon)T .
\]
If $\varepsilon<(\rho-\rho_0)/2$, then $\rho_0+2\varepsilon<\rho$.
Therefore,
\[
    C^\star(\widehat x\mid s,g^\star)<\rho T,
\]
which implies
\[
    \widehat x\in \mathcal{G}_{\rho}(s,g^\star).
\]

\paragraph{Approximate selector.}
The same argument applies when the selector is approximate.
Suppose the learned planner selects $\tilde x\in\mathcal{X}$ such that
\[
    \widehat C(\tilde x\mid s,g^\star)
    \leq
    \min_{x\in\mathcal{X}}\widehat C(x\mid s,g^\star)+\beta T .
\]
Since $x_g\in\mathcal{X}$, we have
\[
    \widehat C(\tilde x\mid s,g^\star)
    \leq
    \widehat C(x_g\mid s,g^\star)+\beta T
    \leq
    (\rho_0+\varepsilon+\beta)T .
\]
Using the uniform error bound once more gives
\[
    C^\star(\tilde x\mid s,g^\star)
    \leq
    \widehat C(\tilde x\mid s,g^\star)+\varepsilon T
    \leq
    (\rho_0+\beta+2\varepsilon)T .
\]
Thus, if $\varepsilon<(\rho-\rho_0-\beta)/2$, then
$C^\star(\tilde x\mid s,g^\star)<\rho T$, and therefore
\[
    \tilde x\in\mathcal{G}_{\rho}(s,g^\star).
\]

\subsection{From Approximate Contraction to Adaptive Stopping}
\label{app:proof_stopping}

Let $T_0$ denote the initial residual horizon.
If each refinement step selects a $\rho$-contraction, then the residual horizon after $L$ refinement steps satisfies
\[
    T_L\leq \rho^L T_0 .
\]
Therefore, to reduce the residual horizon below a local execution radius $r$, it is sufficient to choose
\[
    L
    \geq
    \frac{\log(T_0/r)}{\log(1/\rho)} .
\]

In practice, refinement is learned from finite offline data, so a contraction may fail.
Let $\delta_\rho(T_\ell)$ denote the probability that refinement step $\ell$ fails to select a $\rho$-contraction at residual horizon $T_\ell$.
Let $\psi(T)$ upper-bound the local execution error for a target with residual horizon $T$.

Conditioned on all $L$ refinement steps selecting $\rho$-contractions, the remaining horizon is at most $\rho^L T_0$, so the final local execution error is at most $\psi(\rho^L T_0)$.
The event that the overall procedure fails is contained in the union of the events that at least one refinement step fails and the event that local execution fails after successful refinement.
Therefore, a union bound gives
\[
    P_{\mathrm{err}}(L)
    \leq
    \sum_{\ell=0}^{L-1}\delta_\rho(T_\ell)
    +
    \psi(\rho^L T_0).
\]
This expression captures the refinement trade-off: increasing $L$ can reduce the local execution difficulty, but it also introduces more opportunities for planning errors.
This motivates the value-thresholded stopping rule used by CFHRL: refine while the target is too distant, and stop once the selected target is estimated to be locally executable.
\subsection{Dataset Coverage and Amortized Planner Approximation}
\label{app:coverage-planner}
The analysis in the main text assumes that the candidate set contains a contracting subgoal and that the selector approximately minimizes the estimated decomposition cost.
We now make these two requirements explicit.

For a state--goal pair $(s,g)$ with $T=D^\star(s,g)$, define the dataset contraction mass
\[
q_{\rho_0}(s,g)
=
\Pr_{x\sim \mu_D}
\left[
C^\star(x\mid s,g) \leq \rho_0 T
\right],
\]
where $\mu_D$ denotes the replay-state distribution.
This quantity measures how much probability mass the offline dataset assigns to useful intermediate states.
If $N_c$ candidates are sampled independently from $\mu_D$, then the probability that at least one $\rho_0$-contracting candidate appears is
\[
p_{\mathrm{cov}}(s,g)
=
1-(1-q_{\rho_0}(s,g))^{N_c}.
\]
Thus, increasing the candidate budget improves the chance of observing a replay-supported contracting subgoal, but only when such subgoals have non-negligible dataset support.

\paragraph{Proposition A.1.}
Let $T=D^\star(s,g)$ and suppose $N_c$ candidates are sampled independently from the replay-state distribution $\mu_D$.
Assume that the value-induced decomposition cost satisfies
\[
\sup_{x\in X}
\left|
\widehat C(x\mid s,g)-C^\star(x\mid s,g)
\right|
\leq \epsilon T
\]
with probability at least $1-\delta_V$.
Assume further that the learned planner selects a candidate $\tilde x\in X$ satisfying
\[
\widehat C(\tilde x\mid s,g)
\leq
\min_{x\in X}\widehat C(x\mid s,g)+\beta T
\]
with probability at least $1-\delta_\pi$.
If $\rho_0+\beta+2\epsilon < \rho$, then
\[
\Pr\left[\tilde x\in \mathcal{G}_\rho(s,g)\right]
\geq
1-(1-q_{\rho_0}(s,g))^{N_c}-\delta_V-\delta_\pi.
\]

\paragraph{Proof.}
Let $X$ be the sampled candidate set.
By definition of $q_{\rho_0}(s,g)$, the probability that $X$ contains at least one candidate $x_g$ satisfying
\[
C^\star(x_g\mid s,g)\leq \rho_0T
\]
is
\[
1-(1-q_{\rho_0}(s,g))^{N_c}.
\]
On this coverage event and on the value-accuracy event, we have
\[
\widehat C(x_g\mid s,g)
\leq
C^\star(x_g\mid s,g)+\epsilon T
\leq
(\rho_0+\epsilon)T.
\]
If the planner is $\beta T$-approximate with respect to the estimated cost, then
\[
\widehat C(\tilde x\mid s,g)
\leq
\widehat C(x_g\mid s,g)+\beta T
\leq
(\rho_0+\epsilon+\beta)T.
\]
Applying the value-accuracy bound once more gives
\[
C^\star(\tilde x\mid s,g)
\leq
\widehat C(\tilde x\mid s,g)+\epsilon T
\leq
(\rho_0+\beta+2\epsilon)T.
\]
Since $\rho_0+\beta+2\epsilon<\rho$, we obtain
\[
C^\star(\tilde x\mid s,g)<\rho T,
\]
which implies $\tilde x\in G_\rho(s,g)$.
The stated probability follows by a union bound over the failure of candidate coverage, value accuracy, and planner approximation.

This proposition separates three sources of refinement failure.
The first term corresponds to insufficient replay coverage: the sampled candidate set may not contain any useful intermediate state.
The second term corresponds to error in the learned reachability cost, which may misorder candidates.
The third term corresponds to amortized planner error, since the learned planner may not exactly minimize the estimated decomposition cost.

\section{Implementation Details}
\label{app:implementation}

\subsection{Reachability Score}

We convert the learned value into a non-negative reachability cost by
\begin{equation}
D_\phi(s,g) = \left[-\widehat V_\phi(s,g)\right]_+,
\label{eq:reachability_cost}
\end{equation}
where $[x]_+ = \max(x,0)$. For a single value network, 
$\widehat V_\phi(s,g)=V_\phi(s,g)$. When using an ensemble, we use the conservative estimate
$\widehat V_\phi(s,g)=\min_m V_{\phi_m}(s,g)$.
Since sparse goal-reaching values are non-positive in the ideal case, Eq.~\eqref{eq:reachability_cost}
reduces to $D_\phi(s,g)=-V_\phi(s,g)$, while the clipping only prevents numerical approximation errors
from producing negative reachability costs.

\subsection{Actor Interfaces}
\label{app:actor_interfaces}

CFHRL uses three actor modules with different roles. The high-level actor $\pi_p$ receives the
current observation and the current planning target, and outputs a target in the same state--goal representation used by the reachability model.
The abstraction actor $\pi_z$ receives the current observation and the refined high-level
target, and outputs the command consumed by the low-level actor. The low-level actor $\pi_a$
receives the current observation and this command, and outputs a primitive action.

When learned goal representations are disabled, the abstraction actor outputs a goal vector with the
same dimensionality as the achieved-goal space. When learned representations are enabled, the
abstraction actor outputs a vector in the learned representation space. The sampled representation is
renormalized before being passed to the low-level actor, matching the representation scale used
during training. In the default setting used for the reported experiments, Policy-side learned goal representations are disabled and the mid-level abstraction policy is enabled. All actor losses are implemented as negative log-likelihood losses under diagonal Gaussian policies with constant standard deviations, following the actor interface used by HIQL-style offline GCRL.

\subsection{Offline Candidate Construction}
\label{app:candidate_construction}

The refinement planner is trained using candidate subgoals drawn from the offline dataset. For each high-level training tuple, we sample $N_c$ replay-supported candidate subgoals from the offline dataset. 
Each candidate $\tilde s_i$ is a full dataset observation and is directly used as a candidate target for the planner. 
The decomposition cost is computed as
\[
C_\phi(\tilde s_i\mid s_t,g)
=
\max\{D_\phi(s_t,\tilde s_i),D_\phi(\tilde s_i,g)\}.
\]
Thus, both legs of the decomposition are evaluated by querying the value function with full-observation state-goal pairs. 
This design keeps planner supervision on the support of the offline dataset and avoids introducing an additional goal-representation interface for candidate construction.

\subsection{Training Target Sampling}
\label{app:target_sampling}

The implementation uses different goal-sampling schemes for value learning, low-level actor learning, and high-level planner learning. For value learning, goals are sampled from current, future trajectory, and random goals with probabilities $0.2$, $0.5$, and $0.3$, respectively. For low-level actor learning, actor goals are sampled from future trajectory goals using a fixed-step local-goal sampling strategy. For high-level planner learning, goals are sampled from future trajectory goals and random goals with probabilities $0.8$ and $0.2$, respectively.

For each high-level training tuple, the planner uses $N_c=16$ replay-supported candidate subgoals. This value is used consistently across all reported experiments unless otherwise specified.

\subsection{Training of the Goal Representation}
\label{app:train-goal-rep}
We clarify how the goal representation $\psi_\omega$ is used and trained in
CFHRL. In the reachability/value estimator, the value function is implemented as
\[
    V_{\theta,\omega}(s,g)
    =
    \bar V_\theta\bigl(s, \psi_\omega(s,g)\bigr),
\]
where $\psi_\omega$ is a learned goal encoder. Therefore, $\psi_\omega$ is
trained jointly with the reachability/value estimator through the value loss.
We do not introduce an additional auxiliary representation learning objective
for $\psi_\omega$.

Importantly, $\psi_\omega$ is not updated by the planner, abstraction-policy, or
low-level policy losses. Whenever $\psi_\omega$ is used to construct targets for
these components, the output of $\psi_\omega$ is detached from the computation
graph. For example, the abstraction-policy target is
\[
    z_t =
    \mathrm{sg}\left[
        \psi_\omega(s_t, s_{t+h})
    \right],
\]
where $\mathrm{sg}[\cdot]$ denotes the stop-gradient operator. The abstraction policy is then trained to imitate this detached target using the same weighted negative log-likelihood form as in Eq.~(13):
\[
    \mathcal{L}_{\mathrm{abs}}
    =
    -
    \mathbb{E}
    \left[
        \omega_t
        \log \pi_z\!\left(
        \mathrm{sg}[\psi_\omega(s_t,s_{t+h})]\mid s_t,g
        \right)
    \right].
\]
In our implementation, $\pi_z$ is a diagonal Gaussian actor with a constant standard deviation; thus this loss is equivalent to a weighted squared-error regression up to constants. Gradients from the abstraction-policy loss update only $\pi_z$, not $\psi_\omega$.

\subsection{Inference Schedule}
\label{app:inference_schedule}

At test time, CFHRL repeatedly refines the final goal into a local target and executes the low-level policy for a short interval before replanning. The high-level planner performs at most $K_{\max}$ refinement steps. At each step, the planner samples one subgoal conditioned on the current state and the current planning target. If the current target satisfies $\Dphi(s,g)\leq\epsilon$, refinement stops and the target is passed to the execution module. Otherwise, the proposed subgoal becomes the next planning target. If no target satisfies the reachability threshold within the refinement budget, the last proposed target is used.

The resulting target is passed to the abstraction policy, which produces a local command for the low-level policy. The low-level policy is executed for at most $H_{\mathrm{exec}}$ environment steps before replanning. This schedule keeps the planner adaptive while avoiding expensive test-time search.
\begin{algorithm}[t]
\caption{CFHRL Training}
\label{alg:cfhrl-training}
\begin{algorithmic}[1]
\Require Offline dataset $\mathcal{D}$; value network $\Vphi$; policies $\piplan,\piz,\pia$; candidate count $N_c$.
\Repeat
    \State Sample transitions and hindsight goals from $\mathcal{D}$.
    \State Update $\Vphi$ using $\mathcal{L}_{\mathrm{GCIVL}}$.
    \State Sample high-level tuples $(s_t,g)$ from $\mathcal{D}$.
    \State Construct replay-supported candidate observations 
   $\mathcal{C}_{\mathcal{D}}(s_t,g)=\{\tilde s_i\}_{i=1}^{N_c}$.
    \State Compute 
   $C_\phi(\tilde s_i\mid s_t,g)
   =
   \max\{D_\phi(s_t,\tilde s_i),D_\phi(\tilde s_i,g)\}$.
    \State Compute planner weights $w_i$ by Eq.~\eqref{eq:planner-weight}.
    \State Update $\pi_p$ by weighted imitation of candidate observations $\tilde s_i$.
    \State Sample local segments $(s_t,a_t,s_{t+1},s_{t+h})$ from $\mathcal{D}$ and construct $z_t=\psi(s_t,s_{t+h})$.
    \State Update $\piz$ and $\pia$ using $\mathcal{L}_{\mathrm{z}}$ and $\mathcal{L}_{\mathrm{a}}$.
\Until{convergence}
\end{algorithmic}
\end{algorithm}
\begin{algorithm}[t]
\caption{CFHRL Inference}
\label{alg:cfhrl-inference}
\begin{algorithmic}[1]
\Require Trained $\Dphi,\piplan,\piz,\pia$; current state $s$; final goal $g$; maximum depth $K_{\max}$; threshold $\epsilon$; execution interval $H_{\mathrm{exec}}$.
\While{the episode has not terminated and the final goal has not been reached}
    \State $g^{(0)}\leftarrow g$
    \For{$\ell=0,1,\ldots,K_{\max}-1$}
        \If{$\Dphi(s,g^{(\ell)})\leq \epsilon$}
            \State $\tau\leftarrow \ell$
            \State \textbf{break}
        \EndIf
        \State Sample $u^{(\ell)}\sim \piplan(\cdot\mid s,g^{(\ell)})$
        \State $g^{(\ell+1)}\leftarrow u^{(\ell)}$
        \State $\tau\leftarrow \ell+1$
    \EndFor
    \State $\gbar\leftarrow g^{(\tau)}$
    \For{$j=1,2,\ldots,H_{\mathrm{exec}}$}
        \State Sample $z\sim\piz(\cdot\mid s,\gbar)$
        \State Sample $a\sim\pia(\cdot\mid s,z)$
        \State Execute $a$, observe $s'$, and set $s\leftarrow s'$
        \If{the final goal is reached}
            \State \textbf{break}
        \EndIf
    \EndFor
\EndWhile
\end{algorithmic}
\end{algorithm}
\section{Hyperparameters}
\label{app:hyperparameters}

Table~\ref{tab:hyperparameters} lists the default hyperparameters used in CFHRL. Unless
otherwise stated, these hyperparameters are shared across the reported experiments.

\paragraph{Baseline tuning.}
We use the official OGBench evaluation protocol and official hyperparameter
configurations for all baselines whenever available. Baselines are evaluated on
the same datasets, goal sampling protocol, episode horizons, and number of
random seeds as CFHRL. We do not perform additional per-task tuning for
baselines beyond the official OGBench settings.

\paragraph{Fixed-depth ablations.}
The fixed-depth variants are designed to isolate the contribution of adaptive
stopping. They therefore use the same learned reachability cost $D_\phi$,
planner architecture, abstraction policy, low-level policy, replay data,
optimizer, batch size, training steps, and all other hyperparameters as CFHRL.
The only change is that the recursive refinement procedure is forced to run for
a fixed depth $K$, rather than terminating when
$D_\phi(s, g) \leq \epsilon_{\mathrm{exec}}$.

\paragraph{Environment-specific hyperparameters.}
We report the executability threshold and discount factor used by CFHRL in Table~\ref{tab:env_hparams}. 
The executability threshold $\epsilon_{\mathrm{exec}}$ is used in the adaptive stopping criterion $D_\phi(s,g) \leq \epsilon_{\mathrm{exec}}$, and $\gamma$ is the discount factor used for training the reachability/value estimator. 
Unless otherwise specified, all other hyperparameters are shared across environments and are listed in Table~\ref{tab:hyperparameters}.

\begin{table}[t]
\centering
\caption{
Environment-specific hyperparameters used by CFHRL. All values are fixed before
final evaluation and are not tuned separately on test tasks.
}
\label{tab:env_hparams}
\begin{tabular}{lcccc}
\toprule
Environment family & $\epsilon_{\mathrm{exec}}$ & Discount $\gamma$ & $K_{\max}$ & $H_{\mathrm{exec}}$ \\
\midrule
PointMaze-Giant      & 100 & 0.995 & 3 & 25 \\
AntMaze-Giant        & 100 & 0.995 & 3 & 25 \\
HumanoidMaze-Giant   & 100 & 0.995 & 3 & 100 \\
Cube           & 50 & 0.99 & 1 & 10 \\
Scene          & 50 & 0.99 & 1 & 10 \\
AntSoccer      & 50 & 0.99 & 1 & 25 \\
\bottomrule
\end{tabular}
\end{table}

\begin{table}[t]
\centering
\caption{Default hyperparameters for CFHRL.}
\label{tab:hyperparameters}
\begin{tabular}{lc}
\toprule
Hyperparameter & Value \\
\midrule
Value-learning backend & GCIVL \\
Value ensemble size & 2 \\
Discount $\gamma$ & 0.995 \\
Expectile $\tau$ & 0.9 \\
Optimizer & Adam \\
Learning rate & $3\times 10^{-4}$ \\
Batch size & 256 \\
Target update rate & 0.005 \\
Actor hidden dimensions & $(512,512,512)$ \\
Value hidden dimensions & $(512,512,512)$ \\
Layer normalization & Yes \\
Constant actor standard deviation & Yes \\
Use abstraction policy & Yes \\
Use learned goal representation & No \\
Low-level actor temperature $\alpha_{\mathrm{a}}$ & 10.0 \\
Abstraction actor temperature $\alpha_{\mathrm{z}}$ & 10.0 \\
Advantage weight clip $\omega_{\max}$ & 100.0 \\
Planner temperature $\eta_{\mathrm{p}}$ & 1.0 \\
Training candidate count $N_c$ & 16 \\
Maximum refinement depth $K_{\max}$ & 3 \\
Executability threshold $\epsilon_{\rm exec}$ & 100 \\
Execution interval $H_{\mathrm{exec}}$ & 25 \\
Value loss weight $\lambda_{\mathrm{V}}$ & 1.0 \\
Planner loss weight $\lambda_{\mathrm{p}}$ & 1.0 \\
Abstraction loss weight $\lambda_{\mathrm{z}}$ & 1.0 \\
Low-level actor loss weight $\lambda_{\mathrm{a}}$ & 1.0 \\
Gradient clipping & Disabled \\
\bottomrule
\end{tabular}
\end{table}

\section{Additional Ablation Studies}
\label{app:additional_ablations}

This section provides additional sensitivity analyses for two important inference-time hyperparameters in \ours{}: the maximum number of planning steps and the stopping threshold $\epsilon$ used in the reachability-based refinement rule.
These experiments complement the controlled ablations in Section~\ref{sec:ablations} and further examine how refinement depth and local reachability criteria affect performance.

\subsection{Effect of Planning Steps}
\label{app:planning_steps}

\begin{figure*}[t]
    \centering
    \includegraphics[width=0.92\textwidth]{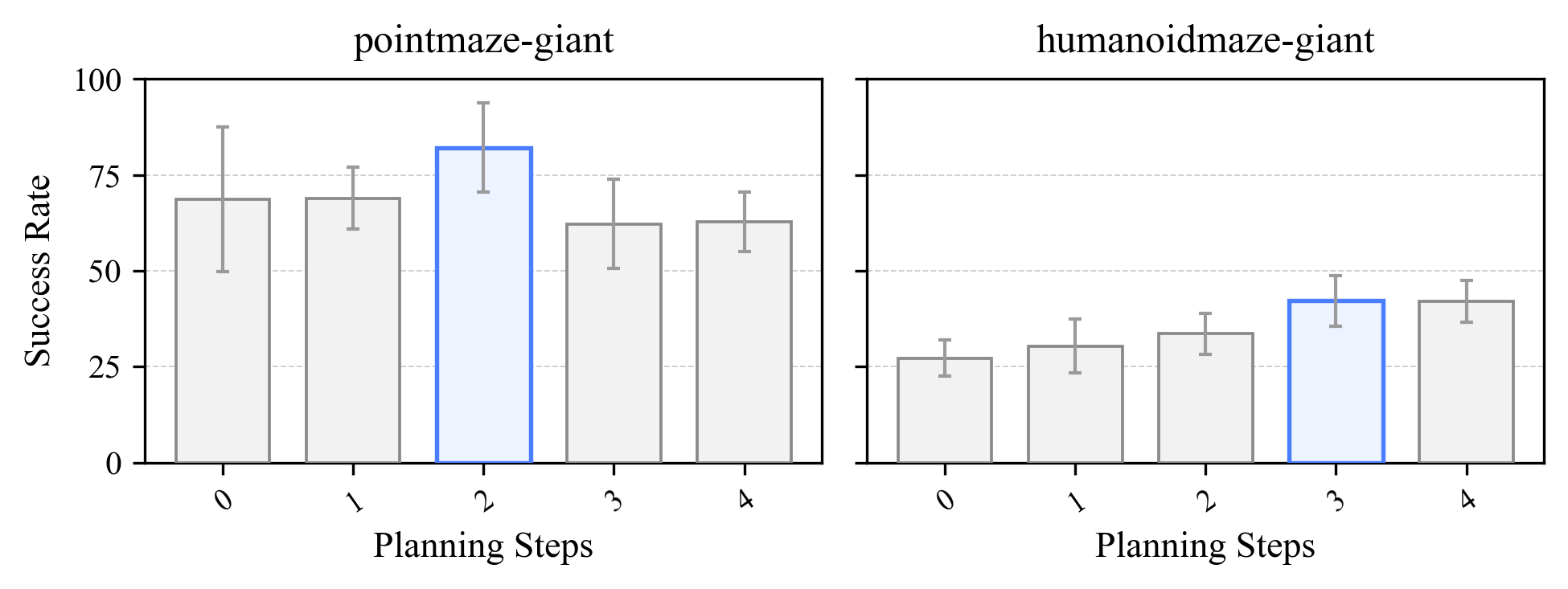}
    \caption{
    Effect of the maximum number of planning steps on PointMaze-Giant and HumanoidMaze-Giant.
    Moderate refinement depth improves performance, while excessive refinement can reduce success due to accumulated planning errors or overly local targets.
    }
    \label{fig:ablation_planning_steps}
\end{figure*}

Figure~\ref{fig:ablation_planning_steps} studies the effect of the maximum number of planning steps.
On PointMaze-Giant, performance improves when the planner is allowed to refine the target for a small number of steps, with the best result obtained around two planning steps.
On HumanoidMaze-Giant, the benefit of additional planning is more gradual, and the best performance is achieved with three to four planning steps.
These results indicate that refinement is useful for long-horizon tasks, but the benefit is not monotonic with planning depth.
Too few planning steps may leave the target outside the reliable execution range, while too many steps can introduce additional planner errors or produce overly conservative intermediate targets.

\subsection{Effect of the Executability Threshold}
\label{app:stopping_threshold}

\begin{figure*}[t]
    \centering
    \includegraphics[width=0.92\textwidth]{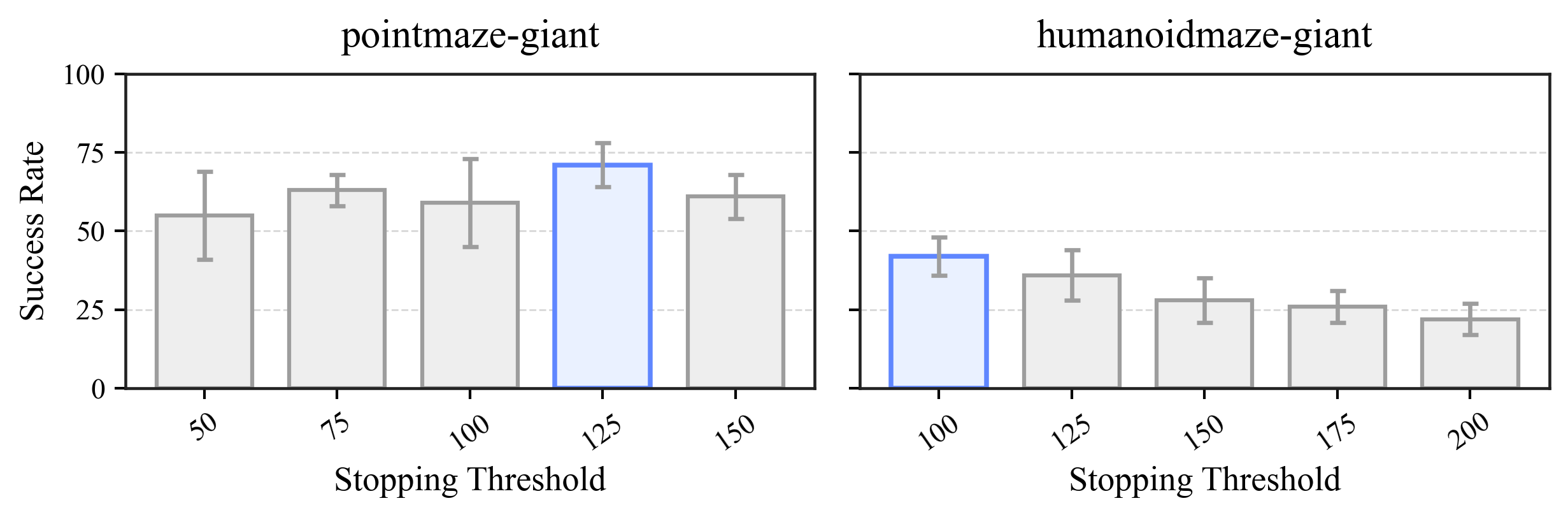}
    \caption{
    Effect of the stopping threshold $\epsilon$ on PointMaze-Giant and HumanoidMaze-Giant.
    The threshold controls when refinement terminates: smaller values require the selected target to be more locally reachable, while larger values stop refinement earlier.
    }
    \label{fig:ablation_stopping_threshold}
\end{figure*}

Figure~\ref{fig:ablation_stopping_threshold} evaluates the sensitivity to the stopping threshold $\epsilon$ in the reachability-based refinement rule.
A smaller threshold imposes a stricter executability criterion and therefore encourages more refinement before execution.
A larger threshold makes the agent stop refinement earlier, but may pass a target to the execution policy before it is reliably reachable.
On PointMaze-Giant, an intermediate threshold achieves the best performance, suggesting that the target should be refined enough to become executable while still preserving meaningful progress.
On HumanoidMaze-Giant, performance is better with a more conservative threshold, indicating that complex dynamics benefit from more local execution targets.
Overall, the results show that the appropriate stopping criterion is environment-dependent, supporting the need for reachability-based adaptive refinement.

\section{Planning Visualization}
\label{app:planning_visualization}
\begin{figure*}[t]
    \centering
    \includegraphics[width=0.92\textwidth]{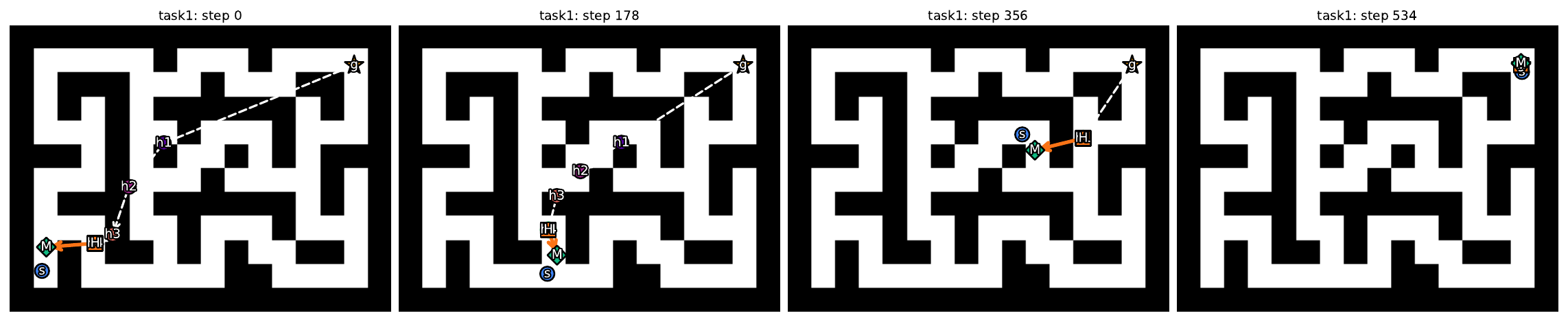}
    \vspace{0.6em}
    \includegraphics[width=0.92\textwidth]{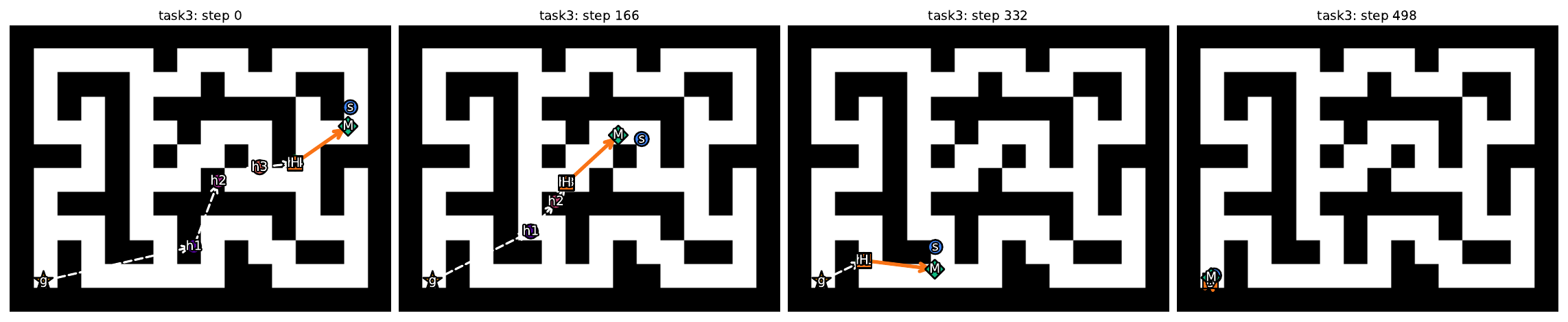}
    \caption{
Visualization of learned coarse-to-fine planning in two representative maze episodes.
Each panel corresponds to a replanning step.
The blue marker $S$ is the current state, the yellow star $g$ is the final goal, and $h_1,h_2,h_3$ are intermediate refinement targets connected by the dashed white line.
The orange square $H$ is the final refined target passed to execution, while the green diamond $M$ is the local command target produced by the abstraction policy.
The orange arrow indicates the resulting local execution direction.
As the agent moves and replans, the selected targets become progressively more local.
}
    \label{fig:planning_visualization}
\end{figure*}

Figure~\ref{fig:planning_visualization} provides qualitative examples of the learned refinement behavior in representative maze episodes.
At each replanning step, \ours{} starts from the final goal and recursively proposes intermediate targets until the selected target is estimated to be executable by the local policy.
The visualization shows that the planner first chooses relatively coarse targets when the final goal is far away, and then produces more local targets as the agent moves closer to the goal.
This behavior is consistent with the intended coarse-to-fine mechanism: intermediate targets are not required to be globally optimal waypoints, but should provide reliable progress and reduce the remaining reaching difficulty.

\section{Limitations}
\label{sec:limitations}

\ours{} has several limitations.
First, its performance depends on the quality and calibration of the learned reachability cost $D_\phi$.
If the value function is severely miscalibrated on off-dataset or contact-rich states, the planner may select poor subgoals.
Second, the current abstraction command is a simple local goal representation in environment-specific goal-relevant coordinates.
This is effective for navigation but can be insufficient for manipulation tasks requiring precise contact sequences.
Third, recursive refinement increases inference-time computation through repeated planner, value, abstraction-policy, and low-level-policy calls, although no replay-buffer candidate enumeration is performed at test time.
Finally, the stylized analysis is illustrative and does not constitute a general theorem for high-dimensional offline RL.
Developing stronger guarantees under realistic value approximation and dataset-coverage assumptions is an important direction for future work.

\paragraph{Generality across value-learning backends.}
Although our implementation uses GCIVL as the default value-learning backend, CFHRL is designed
as a backend-agnostic framework. The planner training objective, the adaptive stopping rule, and
the local execution weights all rely on a generic reachability cost $D_\phi(s,g)$, rather than on
algorithm-specific properties of GCIVL. As a result, the same coarse-to-fine refinement principle can
potentially be instantiated with other offline GCRL value or distance estimators, including GCIQL,
quasimetric representations, contrastive goal-conditioned representations, or conservative value
estimators.

The present paper focuses on validating the refinement mechanism with one stable value backend.
A broader evaluation across different reachability estimators is an important direction for future
work, and would further demonstrate the generality of the proposed framework.

\end{document}